\documentclass{article}
\usepackage{ijcai20}

\usepackage{times}
\usepackage{soul}
\usepackage{url}
\usepackage[hidelinks]{hyperref}
\usepackage[utf8]{inputenc}
\usepackage[small]{caption}
\usepackage{graphicx}
\usepackage{bm}
\usepackage{amsmath}
\usepackage{amsfonts}
\usepackage{amsthm}
\usepackage{booktabs}
\usepackage{subcaption}
\usepackage{algorithm}
\usepackage{algorithmic}
\title{Neural Machine Translation with Error Correction}

\author{
Kaitao Song$^{1}$~\footnote{Corresponding Author: Jianfeng Lu.}
\and
Xu Tan$^2$\And
Jianfeng Lu$^{1}$
\affiliations
$^1$Nanjing University of Science and Technology\\
$^2$Microsoft Research Asia\\
\emails
\{kt.song, lujf\}@njust.edu.cn,
xuta@microsoft.com
}

\newcommand{\ie}{\emph{i.e.}}
\newcommand{\eg}{\emph{e.g.}}

\begin{document}

\maketitle

\begin{abstract}
Neural machine translation (NMT) generates the next target token given as input the previous ground truth target tokens during training while the previous generated target tokens during inference, which causes discrepancy between training and inference as well as error propagation, and affects the translation accuracy. In this paper, we introduce an error correction mechanism into NMT, which corrects the error information in the previous generated tokens to better predict the next token. Specifically, we introduce two-stream self-attention from XLNet into NMT decoder, where the query stream is used to predict the next token, and meanwhile the content stream is used to correct the error information from the previous predicted tokens. We leverage scheduled sampling to simulate the prediction errors during training. Experiments on three IWSLT translation datasets and two WMT translation datasets demonstrate that our method achieves improvements over Transformer baseline and scheduled sampling. Further experimental analyses also verify the effectiveness of our proposed error correction mechanism to improve the translation quality.

\end{abstract}

\section{Introduction}
Neural machine translation (NMT)~\cite{bahdanau2014nmt,Sutskever2014seq2seq,Vaswani2017transformer,song2018dpn,hassan2018achieving} have witnessed great progress due to the development of deep learning. The popular NMT models adopt an encoder-attention-decoder framework, where the decoder generates the target token based on previous tokens in an autoregressive manner. While its popularity, NMT models suffer from discrepancy between training and inference and the consequent error propagation~\cite{Bengio2015SS,Marc2016exposure,wu2019beyond}. During inference, the decoder predicts the next token given previous generated tokens as input, which is discrepant from that in training, where the previous ground-truth tokens as used as input for next token prediction. Consequently, the previous predicted tokens may have errors, which would cause error propagation and affect the prediction of next tokens.

Previous works have tried different methods to solve the above issues, where some of them focus on simulating the data that occurs in inference for training, such as data as demonstration~\cite{Venkatraman2015ImprovingMP}, scheduled sampling~\cite{Bengio2015SS}, sentence-level scheduled sampling~\cite{zhang-etal-2019-bridging}, or even predict them in different directions~\cite{wu2018beyond,tan2019efficient}. While being effective to handle the prediction errors occurred in inference during model training, these methods still leverage the predicted tokens that could be erroneous as the conditional information to predict the next token. Forcing the model to predict correct next token given incorrect previous tokens could be particularly hard and misleading for optimization, and cannot solve the training/inference discrepancy as well as error propagation effectively. 

In this paper, moving beyond scheduled sampling~\cite{Bengio2015SS}, we propose a novel method to enable the model to correct the previous predicted tokens when predicting the next token. By this way, although the decoder may have prediction errors, the model can learn the capability to build correct representations layer by layer based on the error tokens as input, which is more precise for next token prediction than directly relying on previous erroneous tokens as used in scheduled sampling. 

Specifically, we introduce two-stream self-attention, which is designed for language understanding in XLNet~\cite{Yang2019XLNet}, into the NMT decoder to correct the errors while translation. Two-stream self-attention is originally proposed to solve the permutation language modeling, which consists of two self-attention mechanisms: the content stream is exactly the same as normal self-attention in Transformer decoder and is used to build the representations of the previous tokens, while the query stream uses the positional embedding as the inputs to decide the position of the next token to be predicted. In our work, we reinvent two-stream self-attention to support simultaneous correction and translation in NMT, where the content stream is used to correct the previous predicted tokens (correction), and the query stream is used to simultaneously predict the next token with a normal left-to-right order based on the corrected context (translation). 

We conduct experiments on IWSLT 2014 German-English, Spanish-English, Hebrew-English and WMT 2014 English-German and English-Romanian translation datasets to evaluate the effectiveness of our proposed error correction mechanism for NMT. Experimental results demonstrate that our method achieves improvements over Transformer baseline on all tasks. Further experimental analyses also verify the effectiveness of error correction to improve the translation accuracy.

Our contributions can be summarized as follows:
\begin{itemize}
    \item To the best of our knowledge, we are the first to introduce an error correction mechanism during the translation process of NMT, with the help of the newly proposed two-stream self-attention. 
    \item Experimental results on a variety of NMT datasets and further experimental analyses demonstrate that our method achieves improvements over Transformer baseline and scheduled sampling, verifying the effectiveness of our correction mechanism.
\end{itemize}

\section{Background}
In this section, we introduce the background of our work, including the standard encoder-decoder framework, exposure bias and error propagation, and two-stream self-attention mechanism.

\paragraph{Encoder-decoder framework.}
Given a sentence pair $\{x, y\} \in (\mathcal{X}, \mathcal{Y})$, the objective of an NMT model is to maximum the log-likelihood probability ${\rm P}(y|x;\theta)$, where $\theta$ is the parameters of NMT model. The objective function is equivalent to a chain of the conditional probability: ${\rm P}(y|x;\theta) = \prod_{t=1}^n {{\rm P}(y_t|y_{<t},x;\theta)}$, where $n$ is the number of tokens in target sequence $y$ and $y_{<t}$ means the target tokens before position $t$.  The encoder-decoder structure~\cite{Sutskever2014seq2seq,cho2014learning} is the most common framework to solve the NMT task. It adopts an encoder to transform the source sentence $x$ as the contextual information $h$ and a decoder to predict the next token $y_t$ based on the previous target tokens $y_{<t}$ and $h$ autoregressively. Specifically, for the $t$-th token prediction, the decoder feeds the last token $y_{t-1}$ as the input to predict the target token $y_t$. Besides, an encoder-decoder attention mechanism~\cite{bahdanau2014nmt} is used to bridge the connection between source and target sentence.

\begin{figure}[!t]
    \centering
    \begin{subfigure}[t]{0.20\textwidth}
        \centering
        \includegraphics[width=\textwidth]{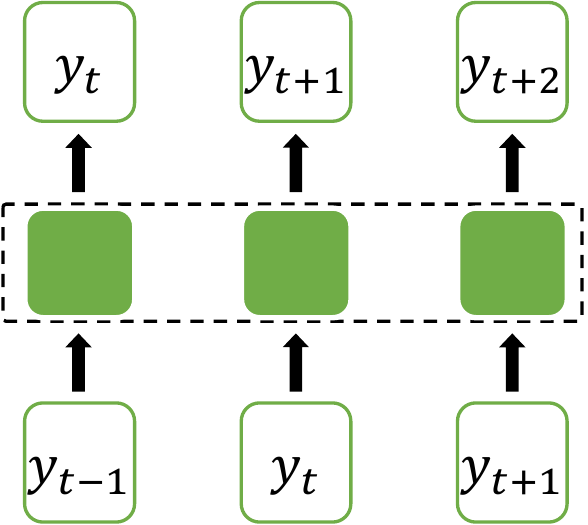}
        \caption{Training}
    \end{subfigure}
    \hspace{0.3cm}
    \begin{subfigure}[t]{0.20\textwidth}
        \centering
        \includegraphics[width=\textwidth]{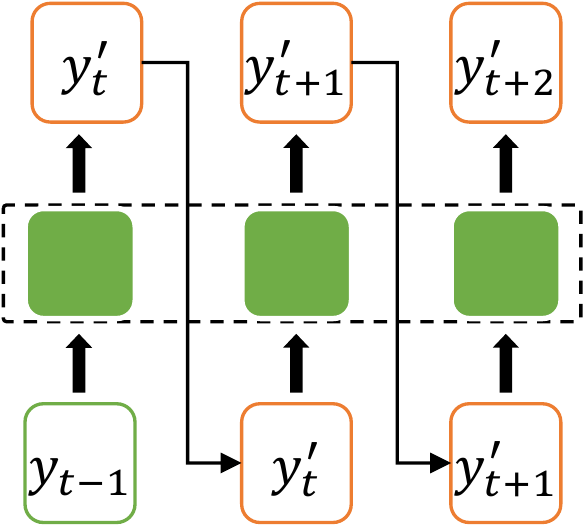}
        \caption{Inference}
    \end{subfigure}
    \caption{The discrepancy between training and inference in autoregressive sequence generation.}
    \label{Exposure_bias}
\end{figure}

\paragraph{Exposure bias and error propagation.} Exposure bias~\cite{Marc2016exposure} is a troublesome problem in the language generation. During the training stage for language generation, it always takes ground truth tokens as the model input. However, at the test stage, the decoder depends on its previous predictions to infer next token. Figure~\ref{Exposure_bias} illustrates us an example of the discrepancy between training and inference in autoregressive sequence generation. 
Once an incorrect token is predicted, the error will accumulate continuously along the inference process. To alleviate this problem, the common solution is to replace some ground truth tokens by predicted tokens at the training time, which is named as \emph{scheduled sampling}. However, scheduled sampling still cannot handle the exposure bias perfectly since it only attempts to predict the next ground truth according to the incorrect predicted tokens but cannot reduce the negative effect from incorrect tokens. Therefore, we propose a novel correction mechanism during translation to alleviate the error propagation.

\paragraph{Two-stream self-attention.}
XLNet is one of the famous pre-trained methods~\cite{radford2019language,devlin2019bert,Yang2019XLNet,song2019mass} for natural language processing.
It first proposed a two-stream self-attention mechanism, which consists of a content stream and a query stream for permutation language modeling. For token $y_t$, it can see tokens $y_{\le t}$ in the content stream while only see tokens $y_{<t}$ in the query stream. Beneficial from two-stream self-attention mechanism, the model can predict the next token in any position with the corresponding position embedding as the query, in order to enable permutation language modeling. Besides, Two-stream self-attention also avoids the pre-trained model to use $\rm{[MASK]}$ token into the conditioned part during the training, where the $\rm{[MASK]}$ token would bring mismatch between pre-training and fine-tuning. In this paper, we leverage the advantages of two-stream self-attention to design a novel error correction mechanism for NMT.

\begin{figure}[!t]
    \centering
    \begin{subfigure}[t]{0.49\textwidth}
        \centering
        \includegraphics[width=\textwidth]{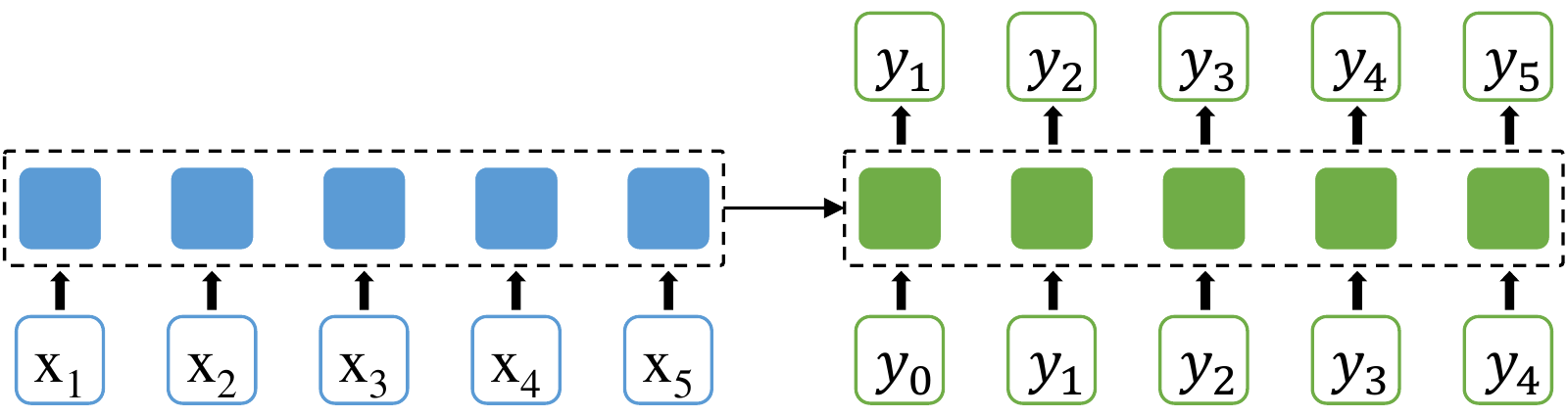}
        \caption{Standard NMT}
        \label{gpt}
    \end{subfigure}
    \vspace{5pt}
    \begin{subfigure}[t]{0.49\textwidth}
        \centering
        \includegraphics[width=\textwidth]{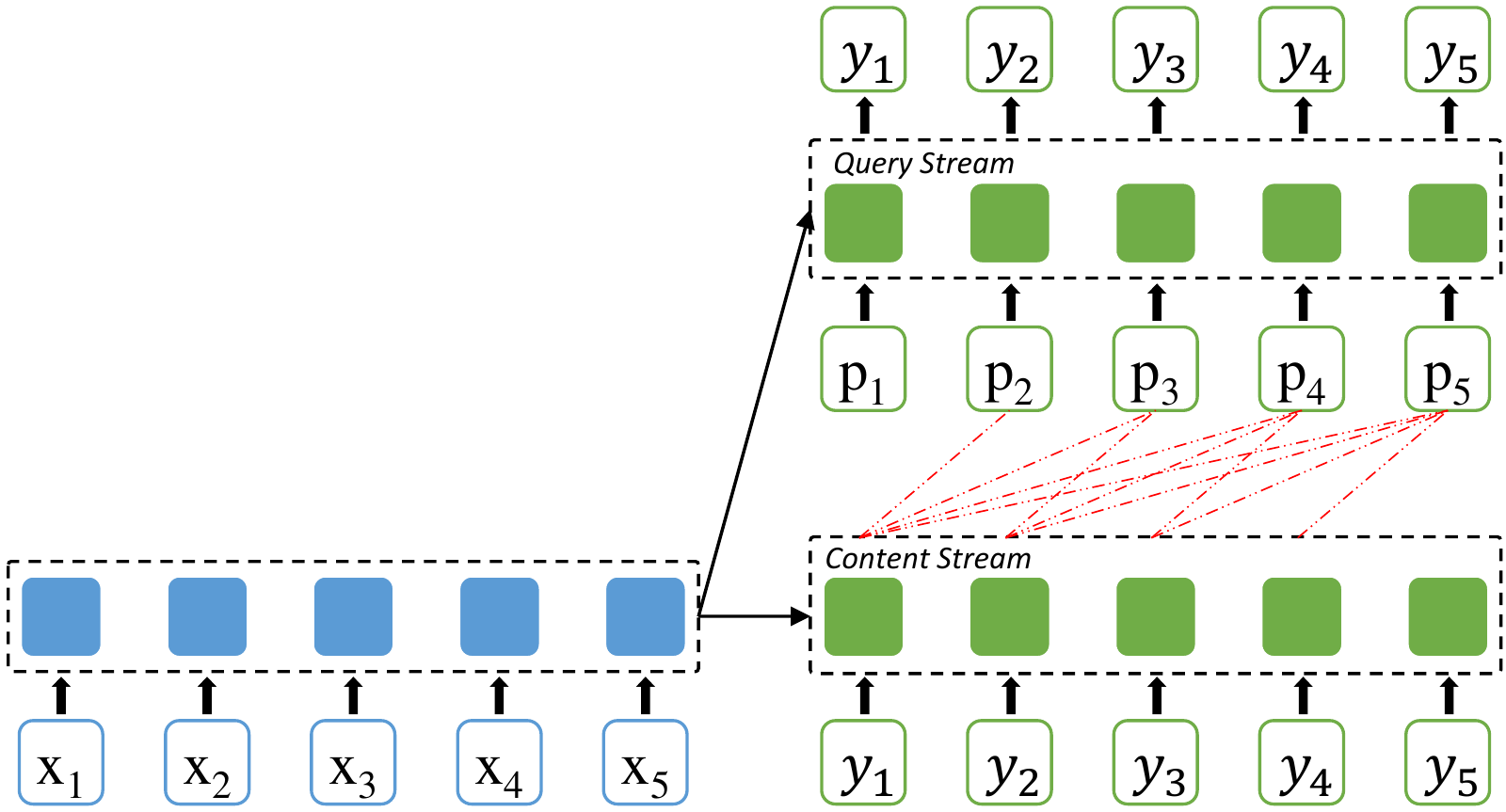}
        \caption{NMT with two-stream self-attention. $p_i$ means the position of $i$-th token. The red dashed line means that the query stream can attend to the content stream.}
        \label{tssa}
    \end{subfigure}
    \vspace{5pt}
    \begin{subfigure}[t]{0.49\textwidth}
        \centering
        \includegraphics[width=\textwidth]{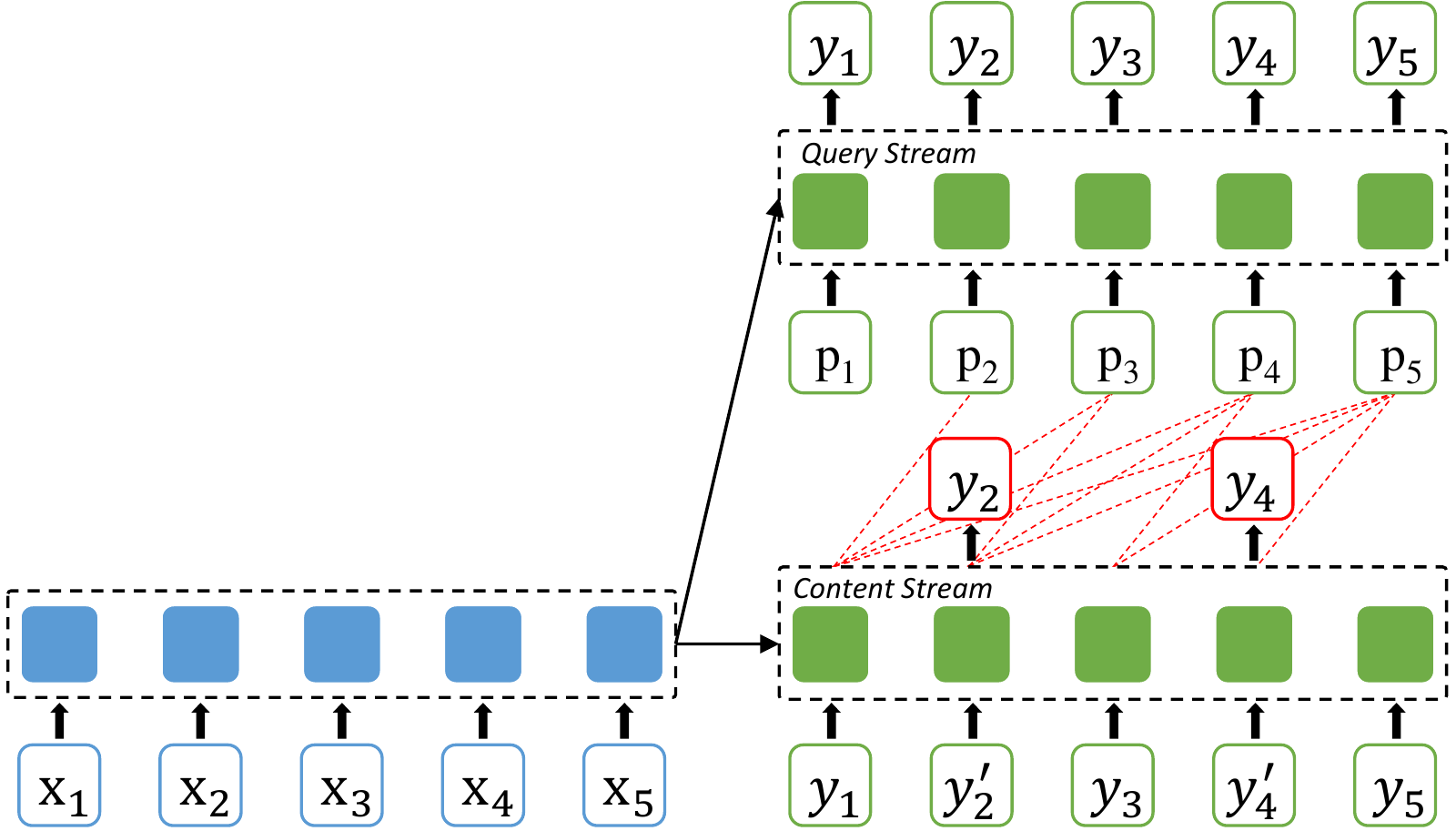}
        \caption{NMT with our proposed error correction mechanism based on two-stream self-attention. $y'_i$ is predicted tokens by the model itself. The cells in red color represent that the potential error token $y'_2$ and $y'_4$ are corrected into the ground truth tokens $y_2$ and $y_4$.}
        \label{correct}
    \end{subfigure}
    \caption{The illustrations of standard NMT, NMT with two-stream self-attention and our proposed error correction mechanism on two-stream self-attention.}
    \label{NMT}
\end{figure}

\section{Method}

Previous NMT models~\cite{bahdanau2014nmt,Vaswani2017transformer} generate the next tokens $y'_t$ from the probability (\ie, $y'_t \sim {\rm P}(y_t|y_{<t},x;\theta)$). If $y'_t$ is predicted with error and taken as the next decoder input, can we force model to automatically build correct hidden representations that are close to the ground truth token $y_t$. In this case, the consequent token generation can build upon the previous correct representations and become more precise. A natural idea is to optimize the model to maximize the correction probability ${\rm P}(y_t|y'_t, y_{<t},x;\theta)$ simultaneously when maximizing the probability of next token prediction ${\rm P}(y_{t+1}|y'_t,y_{<t},x;\theta)$. However, the previous NMT decoder does not well support this correction mechanism. Inspired by the two-stream self-attention in XLNet~\cite{Yang2019XLNet}, we leverage the content stream to maximize ${\rm P}(y_t|y'_t, y_{<t},x;\theta)$ and the query stream to maximize ${\rm P}(y_{t+1}|y'_t,y_{<t},x;\theta)$, which can well meet the requirements of simultaneous correction and translation. 

In order to introduce our method more clearly, in the following subsections, we first introduce the integration of two-stream self-attention into NMT model, and then introduce the error correction mechanism for NMT based on two-stream self-attention. 

\subsection{NMT with Two-Stream Self-Attention}
 
Inspired by XLNet~\cite{Yang2019XLNet},
we incorporate the idea of two-stream self-attention to modify the decoding of NMT framework.  
Specifically, the encoder is the same as the standard NMT model, and the decoder is incorporated with two-stream self-attention where the positional embedding is taken as input in query stream for prediction while content stream is used to build context representations. Different from that in XLNet, we make two modifications: 1) we remove the permutation language modeling in XLNet since the decoder in NMT usually only uses left-to-right generation, and 2) we let the decoder to predict the whole sequence rather than partial sentence in XLNet. Figure~\ref{gpt} and Figure~\ref{tssa} show the differences between the standard NMT and the NMT with two-stream self-attention.

We formulate the NMT with two-stream self-attention as follows. For the decoder, we feed the positions $\{p_1, \cdots, p_n\}$ to the query stream to provide the position information for the next token prediction, and the sequence $\{y_1, \cdots, y_n\}$ plus its positions $\{p_1, \cdots, p_n\}$ to the content stream to build contextual information. For the $l$-th layer, we define the hidden states of query/content streams as $q_t^l$ and $c_t^l$. The updates for the query and content streams are as follows:
\begin{align}
    {q_t^{l+1}} &= {\rm Attention}({\rm Q}=q_t^l, {\rm KV}=c_{< t}^{l};h,\theta_{l+1}) \\
    {c_t^{l+1}} &= {\rm Attention}({\rm Q}=c_t^l, {\rm KV}=c_{\le t}^{l};h,\theta_{l+1}),
\end{align}
where $h$ represents the hidden states from encoder outputs and $\theta_{l+1}$ represents the parameters of the ${l+1}$ layer, $\rm{Q}$ and $\rm{KV}$ represents the query, key and value in self-attention. Both the query and content stream share the same model parameters. The states of key and value can be reused in both query and content streams. Finally, we feed the outputs of the query stream from the last layer to calculate the log-probability for the next target-token prediction. During inference, we first predict the next token with the query stream, and then update the content stream with the generated token. The order of query and content stream will not affect the predictions since the tokens in the query stream only depend on the previously generated tokens of the content streams.

\subsection{Error Correction based on Two-Stream Self-Attention}
\label{sec3_3}

Benefiting from the two-stream self-attention, we can naturally introduce an error correction mechanism on the content stream. The content stream is originally designed to build the representations of previous tokens, which is used in query stream for next token prediction. In order to correct errors, the content stream also needs to predict the correct tokens given incorrect tokens as input. 

In order to simulate the prediction errors in the input of the content stream, we also leverage scheduled sampling~\cite{Bengio2015SS} to randomly sample tokens either from the ground truth $y=\{y_1,\cdots,y_n\}$ or the previously predicted tokens $y'=\{y'_1,\cdots,y'_n\}$ with a certain probability as the new inputs $\Tilde{y}=\{\Tilde{y}_1,\cdots,\Tilde{y}_n\}$, where $y'_t$ is sampled from the probability distribution ${\rm P}(y_t|y_{<t},x;\theta)$. For inputs $\Tilde{y}_t$, it equals to $y_t$ with a probability $p(\cdot)$ otherwise $y'_t$. For each token $y'_t$ ($y'_t \ne y_t$) predicted by the query stream in step $t$, we force the content stream to predict its corresponding ground truth token $y_t$ again. The loss function for the error correction mechanism (ECM) is formulated as:
\begin{equation}
\label{eq4}
    \mathcal{L}_{{\rm ECM}}(y|\Tilde{y},x;\theta) = -\sum_{t=1}^n 1(\Tilde{y}_{t} \ne y_t)\log({\rm P}(y_t|\Tilde{y}_{\le t},x;\theta)).
\end{equation}
In the mechanism, the content stream can learn to gradually correct the hidden representations of error tokens toward the correct counterpart layer by layer. 

The query stream is still used to predict the the next token, given a random mixture of previous predicted tokens and ground truth tokens. The negative log-likelihood (NLL) loss is formulated as:
\begin{equation}
\label{eq3}
     \mathcal{L}_{{\rm NLL}}(y|\Tilde{y},x;\theta) = -\sum_{t=1}^n \log({\rm P}(y_t|\Tilde{y}_{<t},x;\theta)).
\end{equation}
Finally, we combine the two loss functions as the final objective function for our method:
\begin{equation}
    \min \mathcal{L}_{\rm{NLL}}(y|\Tilde{y},x;\theta) + \lambda \cdot \mathcal{L}_{{\rm ECM}}(y|\Tilde{y},x;\theta),
\end{equation}
where $\lambda$ is a hyperparameter to balance the NLL loss and ECM loss. 

Figure~\ref{correct} demonstrates the workflow of our proposed error correction mechanism. The difference between our error correction mechanism and the naive scheduled sampling is that once an error token is predicted in scheduled sampling, the model still learns to predict the next correct token given error tokens as context, which could confuse the the model and mislead to learn incorrect prediction patterns. However, based on our error correction mechanism, the next token prediction is built upon the representations that are corrected by the content stream, and is more precise to learn prediction patterns.

In our error correction mechanism, how to control the scheduled sampling probability $p(\cdot)$ and when to sample tokens are important factors for the training. Previous works~\cite{Bengio2015SS} indicated that it is unsuitable to sample tokens from scratch during the training since the model is still under-fitting and the sampled tokens will be too erroneous. Inspired by OR-NMT~\cite{zhang-etal-2019-bridging}, we design a similar exponential decay function for sampling probability $p(\cdot)$ but with more restrictions. The decay function is set as:

\begin{equation}
p(s) = 
    \begin{cases}
        1, & s \le \alpha \\
        \max(\beta, \frac{\mu}{\mu + \exp((s-\alpha)/\mu)}), & {\rm otherwise} \\
    \end{cases},
\end{equation}
where $s$ represents the training step, $\alpha$, $\beta$ and $\mu$ are hyperparameters. The hyperparameter $\alpha$ means the step when model starts to sample tokens, and hyperparameter $\beta$ is the maximum probability for sampling. 

\section{Experimental Setting}
In this section, we introduce the experimental settings to evaluate our proposed method, including datasets, model configuration, training and evaluation.
\subsection{Datasets}
We conduct experiments on three IWSLT translation datasets (\{German, Spanish, Hebrew\} $\rightarrow$ English) and two WMT translation datasets (English $\rightarrow$ \{German, Romanian\}) to evaluate our method. In the follow sections, we abbreviate English, German, Spanish, Hebrew, Romanian as ``En", ``De", ``Es", ``He", ``Ro".
\paragraph{IWSLT datasets.} For IWSLT14 De$\to$En, it contains 160K and 7K sentence pairs in training set and valid set. We concatenate TED.tst2010, TED.tst2011, TED.tst2012, TED.dev2010 and TED.tst2012 as the test set. For IWLST14 Es$\to$En and He$\to$En~\footnote{IWSLT datasets can be download from \url{https://
wit3.fbk.eu/archive/2014-01/texts}}, they contain 180K and 150K bilingual data for training. We choose TED.tst2013 as the valid set and TED.tst2014 as the test set. During the data preprocess, we learn a 10K byte-pair-coding (BPE)~\cite{sennrich2016bpe} to handle the vocabulary.
\paragraph{WMT datasets.} WMT14 En$\to$De and WMT16 En$\to$Ro translation tasks contain 4.5M and 2.8M bilingual data for training. Following previous work~\cite{Vaswani2017transformer}, we concatenate newstest2012 and newstest2013 as the valid set, and choose newstest2014 as the test set for WMT14 En$\to$De. For WMT16 En$\to$Ro, we choose newsdev2016 as the valid set and newstest2016 as the test set. We learn 32K and 40K BPE codes to tokenize WMT14 En$\to$De and WMT16 En$\to$Ro dataset.

\subsection{Model Configuration}
We choose the state-of-the-art Transformer~\cite{Vaswani2017transformer} as the default model. For IWSLT tasks, we use 6 Transformer blocks, where attention heads, hidden size and filter size are 4, 512 and 1024. Dropout is set as 0.3 for IWSLT tasks. The parameter size is 39M. For WMT tasks, we use 6 Transformer blocks, where attention heads, hidden size and filter size are 16, 1024 and 4096. And the parameter size is as 214M. Dropout is set as 0.3 and 0.2 for En$\to$De and En$\to$Ro respectively. To make a fair comparison, we also list some results by the original NMT model, without two-stream self-attention. For the decay function of sampling probability, we set $a$, $b$ and $\mu$ as 30,000, 0.85 and 5,000. The $\lambda$ for $\mathcal{L}_{{\rm ECM}}$ is tuned on the valid set, and a optimal choice is 1.0. To manifest the advances of our method, we also prepare some strong baselines for reference, including: Layer-wise Transformer~\cite{he2018layer_wise}, MIXER~\cite{Marc2016exposure} on Transformer and Tied-Transformer~\cite{Yingce2019tied}.
\subsection{Training and Evaluation}
During training, we use Adam~\cite{kingma2014method} as the default optimizer, with a linear decay of learning rate. The IWSLT tasks are trained on single NVIDIA P40 GPU for 100K steps and the WMT tasks are trained with 8 NVIDIA P40 GPUs for 300K steps, where each GPU is filled with 4096 tokens.
During inference, We use beam search to decode results. The beam size and length penalty is set as 5 and 1.0 for each task except WMT14 En$\to$De, which use a beam size of 4 and length penalty is 0.6 by following the previous work~\cite{Vaswani2017transformer}. All of the results are reported by multi-bleu~\footnote{\url{https://github.com/moses-smt/mosesdecoder/blob/master/scripts/generic/multi-bleu.perl}}. 
Our code is implemented on fairseq~\cite{ott-etal-2019-fairseq}~\footnote{https://github.com/pytorch/fairseq}, and we will release our code under this link: \url{https://github.com/StillKeepTry/ECM-NMT}.
	
\begin{table}[h]
	\centering
	\begin{tabular}{l|c c c}
		\toprule
		Method & De$\to$En & Es$\to$En & He$\to$En \\
		\midrule
		Tied Transformer & 35.10 & 40.51 & \\
		Layer-Wise Transformer & 35.07 & 40.50 & - \\
		MIXER   & 35.30 & 42.30 & - \\
		\midrule
		Transformer Baseline & 34.78 & 41.78 & 35.32 \\
		Our method &  \textbf{35.70} & \textbf{43.05} & \textbf{36.49} \\
		\bottomrule
	\end{tabular}
	\caption{BLEU score on IWSLT14 Translation tasks in different setting.}
	\label{IWSLT}
\end{table}

\section{Results}
In this section, we report our result on three IWSLT tasks and two WMT tasks. Furthermore, we also study each hyperparameter used in our model, and conduct ablation study to evaluate our method.

\subsection{Results on IWSLT14 Translation Tasks}
The results of IWLST14 tasks are reported in Table~\ref{IWSLT}. 
From Table~\ref{IWSLT}, we find our model with correction mechanism outperforms baseline by 0.89, 0.99 and 0.83 points and the original NMT baseline by 0.92, 1.27 and 1.17 points on De$\to$En, Es$\to$En and He$\to$En respectively. Note that our baseline is strong enough which is comparable to the current advanced systems. Even within such strong baselines, our method stills achieves consistent improvements in all three tasks. These improvements also confirm the effectiveness of our method in correcting error information. 
	
\begin{table}[!t]
	\centering
	\begin{tabular}{l|c|c}
		\toprule
		Method & En$\to$De & En$\to$Ro \\
		\midrule
		Tied Transformer & 28.98 & 34.67 \\
		Layer-wise Transformer & 29.01 & 34.43 \\
		MIXER   & 28.68 & 34.10 \\
		\midrule
		Transformer Baseline & 28.40 & 32.90 \\
		Our method       & \textbf{29.20} & \textbf{34.70} \\
		\bottomrule
	\end{tabular}
	\caption{BLEU score on WMT14 En$\to$De and WMT16 En$\to$Ro.}
	\label{WMT}
\end{table}

\subsection{Results on WMT Translation Tasks}
In order to validate the performance of our method on large-scale datasets, we also conduct experiments on WMT14 En$\to$De and WMT16 En$\to$Ro. The results are reported in Table~\ref{WMT}. 
We found when incorporating error correction mechanism into the NMT model, it can achieve 29.20 and 34.70 BLEU score, which outperforms our baseline by 0.8 and 1.6 points in En$\to$De and En$\to$Ro, and is comparable to the previous works. These significant improvements on two large-scale datasets also demonstrate the effectiveness and robustness of our method in solving exposure bias. In addition, our approach is also compatible with previous works, \ie, our method can achieve better performance if combined with other advanced structure. 

\begin{table}[h]
	\centering
	\begin{tabular}{l|c|c|c}
		\toprule
		Method & De$\to$En & Es$\to$En &  En$\to$De \\
		\midrule
		Our method & \textbf{35.70} & \textbf{43.05} & \textbf{29.20} \\
		\quad -ECM  & 35.40  & 42.55  & 28.72 \\
		\quad -ECM -SS  & 34.81  & 42.16  & 28.48 \\
	    \quad -ECM -SS -TSSA & 34.78  & 41.81  & 28.40 \\
		\bottomrule
	\end{tabular}
	\caption{Ablation study of different component in our model. The second, third and forth column are results on IWSLT14 De$\to$En, Es$\to$En and WMT14 En$\to$De. The second row is our method. The third row is equal to the second row removing error correction mechanism (ECM). The forth row is equal to the third row removing scheduled sampling (SS). The last raw is equal to the forth row removing two-stream self-attention, \ie, the standard NMT. The prefix ``-" means removing this part. }
	\label{SS_study}
\end{table}

\subsection{Ablation Study}
To demonstrate the necessity of each component in our method, we take a series of ablation study on our model on IWSLT14 De$\to$En, Es$\to$En and WMT14 En$\to$De. The results are shown in Table~\ref{SS_study}. When disabling error correction mechanism (ECM), we find the model accuracy decreases 0.30, 0.48 and 0.50 points respectively in different tasks. When further removing scheduled sampling (SS), the model accuracy drop to 34.81, 41.16, 28.48 points in three tasks. We observe that in the large-scale dataset, the improvements for scheduled sampling are limited, while our model still achieves stable improvements in the large-scale dataset, which proves the effectiveness of the error correction mechanism. 
In addition, we also make a comparison between the original NMT and the NMT with two-stream self-attention (TSSA) to verify whether two-stream self-attention mechanism will contribute to model accuracy improvement. From Table~\ref{SS_study}, we find the NMT model with TSSA performs slightly better than the original NMT model in the small-scale tasks by 0.03-0.35 points and is close to the accuracy of the original NMT model in the large-scale tasks. This phenomenon also explains the improvement of our method is mainly brought by the error correction, rather than the two-stream self-attention. In summary, every component all plays an indispensable role in our model.

\begin{figure}[b]
    \centering
    \begin{subfigure}[t]{0.235\textwidth}
        \centering
        \includegraphics[width=\textwidth]{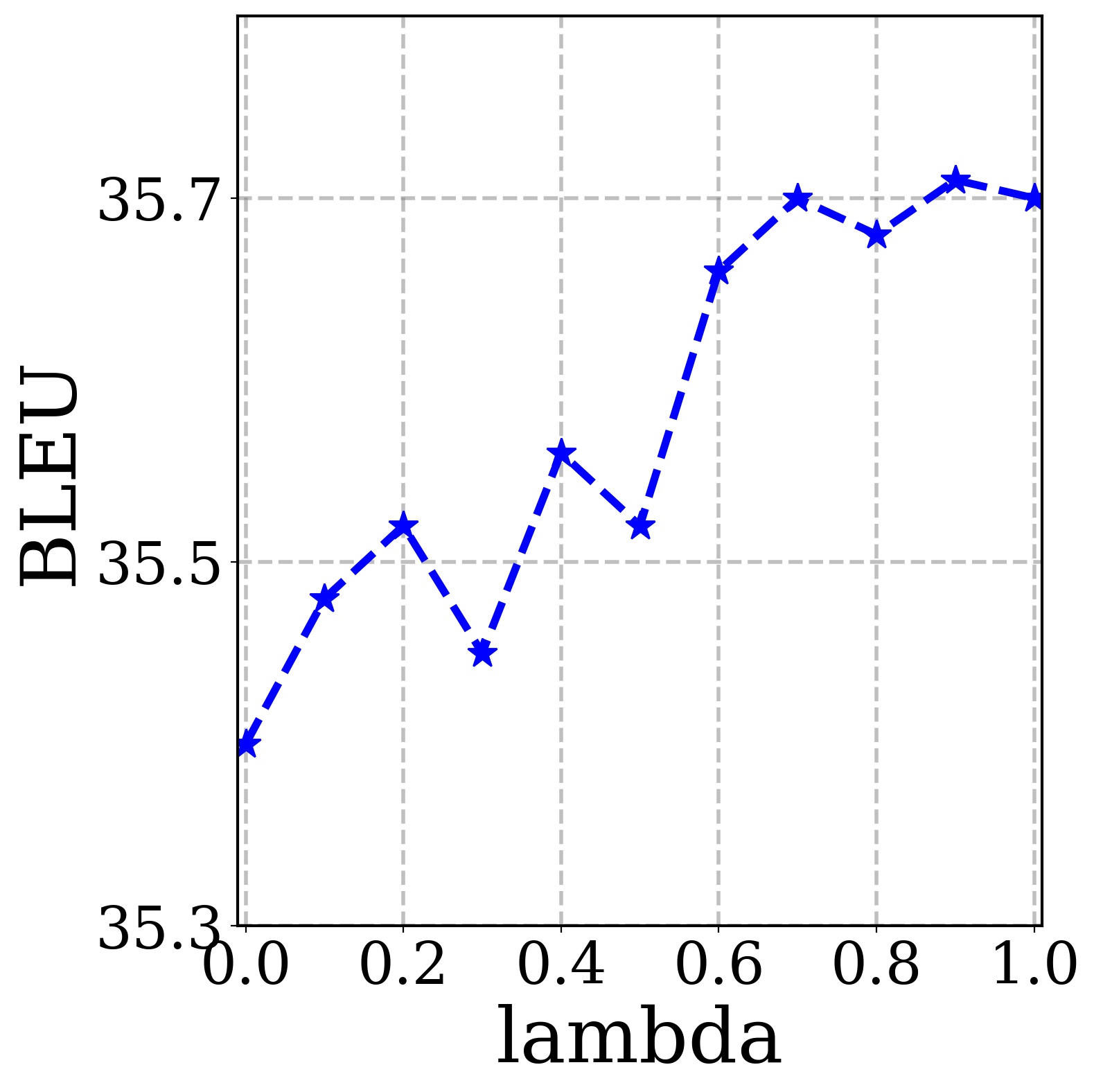}
        \caption{IWSLT14 De$\to$En}
        \label{iwslt14_lambda}
    \end{subfigure}
    \begin{subfigure}[t]{0.235\textwidth}
        \centering
        \includegraphics[width=\textwidth]{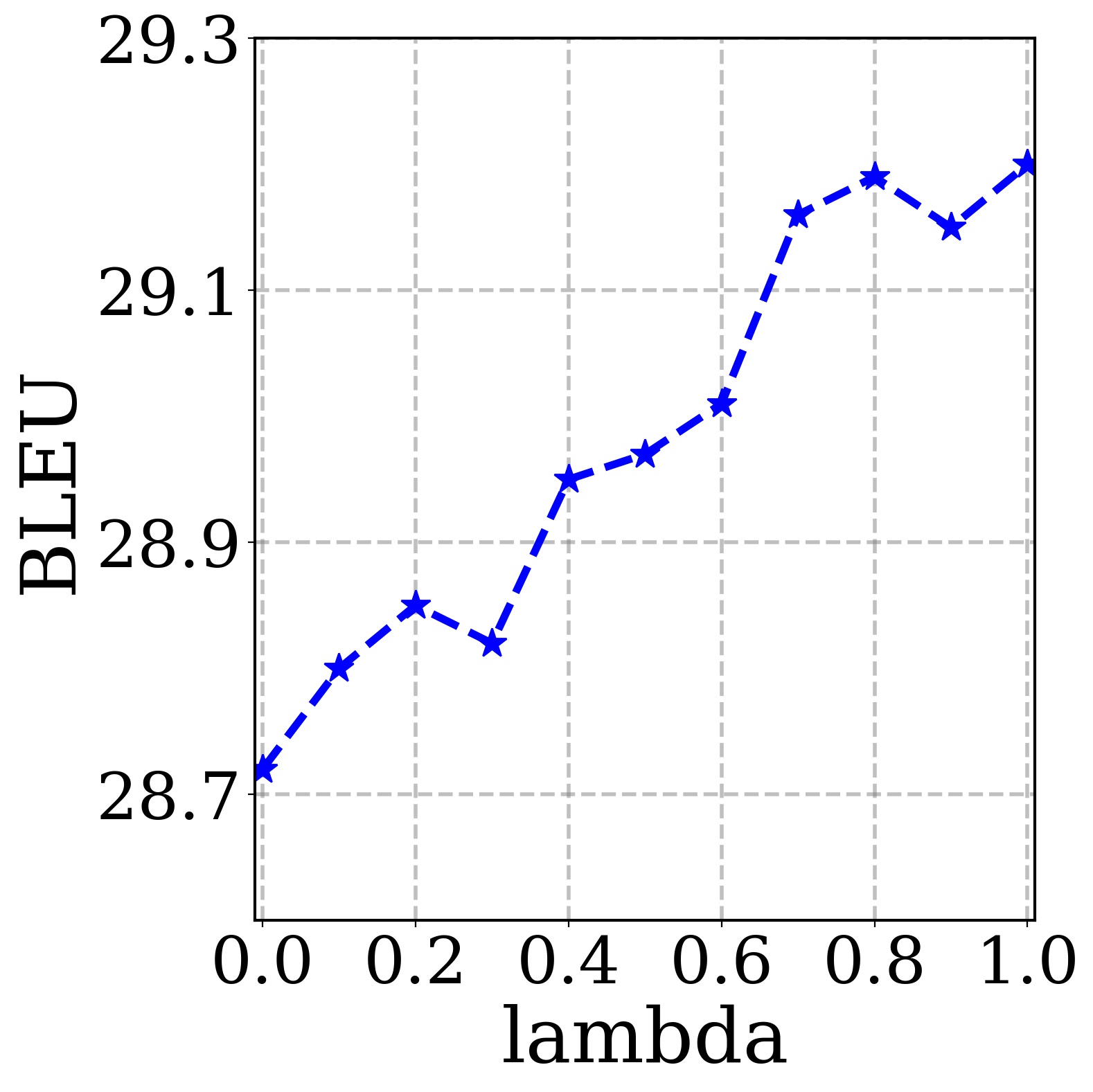}
        \caption{WMT14 En$\to$De}
        \label{wmt14_ende}
    \end{subfigure}
    \caption{Results on IWSLT14 De$\to$En and WMT14 En$\to$De with different $\lambda$.}
    \label{lambda_study}
\end{figure}

\begin{table*}[!t]
	\centering
	\begin{tabular}{l r l}
		\toprule
		Method & & Translation \\
		\cmidrule{1-1} \cmidrule{3-3}
		Source (De) &  & in dem moment war es , als ob ein filmregisseur einen bühnenwechsel verlangt hätte . \\
		\cmidrule{1-1} \cmidrule{3-3}
		Target (En) & & at that moment , it was as if a film director called for a set change . \\
		\cmidrule{1-1} \cmidrule{3-3}
		Baseline & & it was \emph{like a movie} director \emph{had demanded a shift} at the moment . \\
		\cmidrule{1-1} \cmidrule{3-3}
		Baseline + SS & & at \emph{the} moment , it was \emph{like} a \emph{filmmaker had requested} a change. \\
		\cmidrule{1-1} \cmidrule{3-3}
		Our method & & at \emph{the} moment , it was as if a \emph{movie} director \emph{had} called for a change. \\
		\bottomrule
	\end{tabular}
	\vspace{3pt}
	\caption{A translation case on IWSLT14 De$\to$En test set, generated by the baseline method, baseline with scheduled sampling and our method with error correction. The italic font means the mismatch translation.}
	\label{de-en_cases}
\end{table*}

\subsection{Study of $\lambda$ for $\mathcal{L}_{{\rm ECM}}$}
To investigate the effect of $\lambda$ in $\mathcal{L}_{{\rm ECM}}$ on model accuracy, we conduct a series of experiments with different $\lambda$ on IWSLT14 De$\to$En and WMT14 En$\to$De datasets. The results are shown in Figure~\ref{lambda_study}. It can be seen that the accuracy improves with the growth of $\lambda$. When $\lambda \ge 0.6$, the model accuracy increases slowly. The model achieves best accuracy at $\lambda = 0.9$ in IWSLT14 De$\to$En and $\lambda = 1.0$ in WMT14 En$\to$De.
To avoid the heavy cost in tuning hyperparameters for different tasks, we limit $\lambda$ as 1.0 in all of final experiments. 

\begin{figure}[!t]
    \centering
    \begin{subfigure}[t]{0.237\textwidth}
        \centering
        \includegraphics[width=\textwidth]{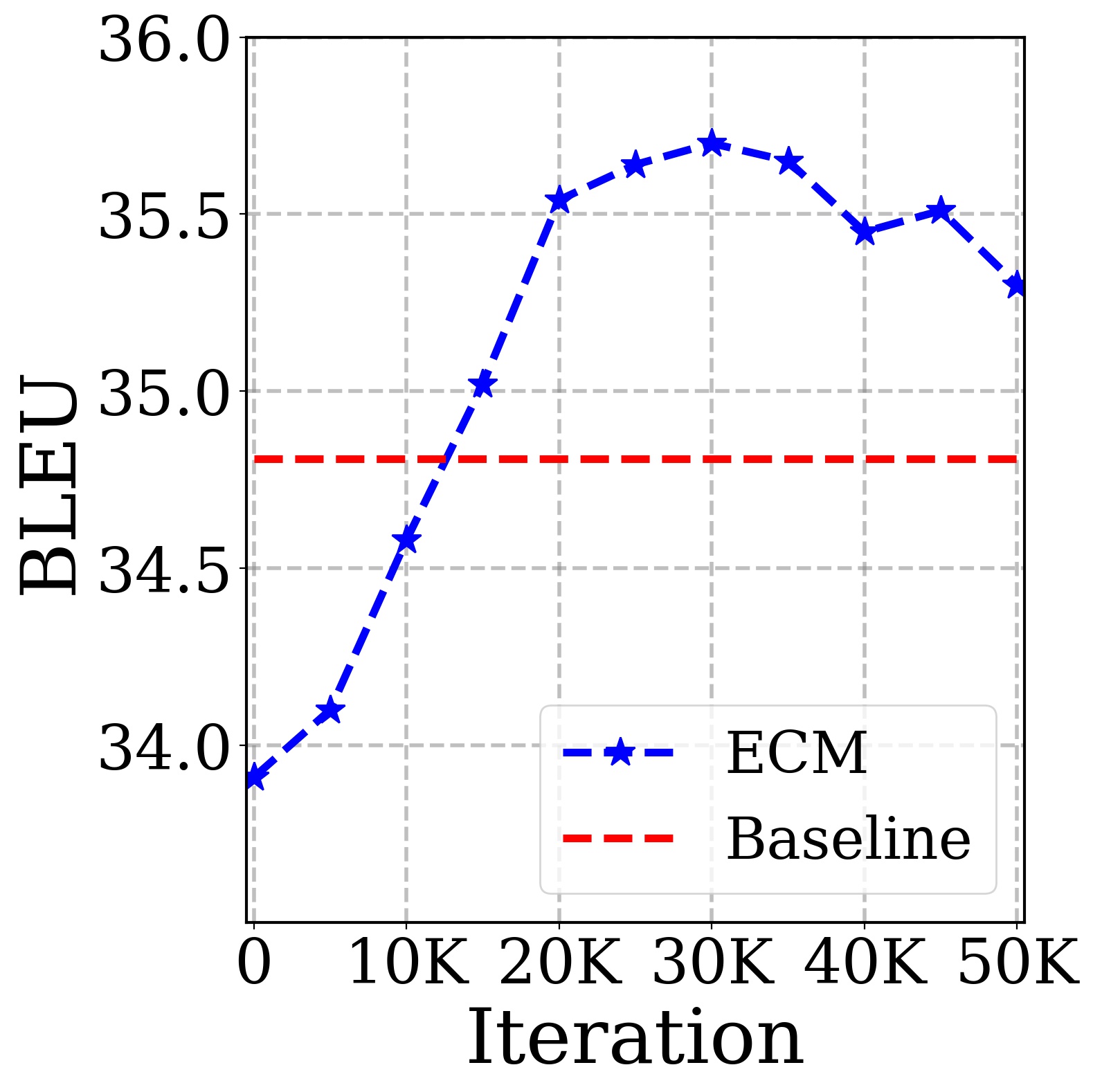}
        \caption{Study of $\alpha$}
        \label{hyper_a}
    \end{subfigure}
    \begin{subfigure}[t]{0.237\textwidth}
        \centering
        \includegraphics[width=\textwidth]{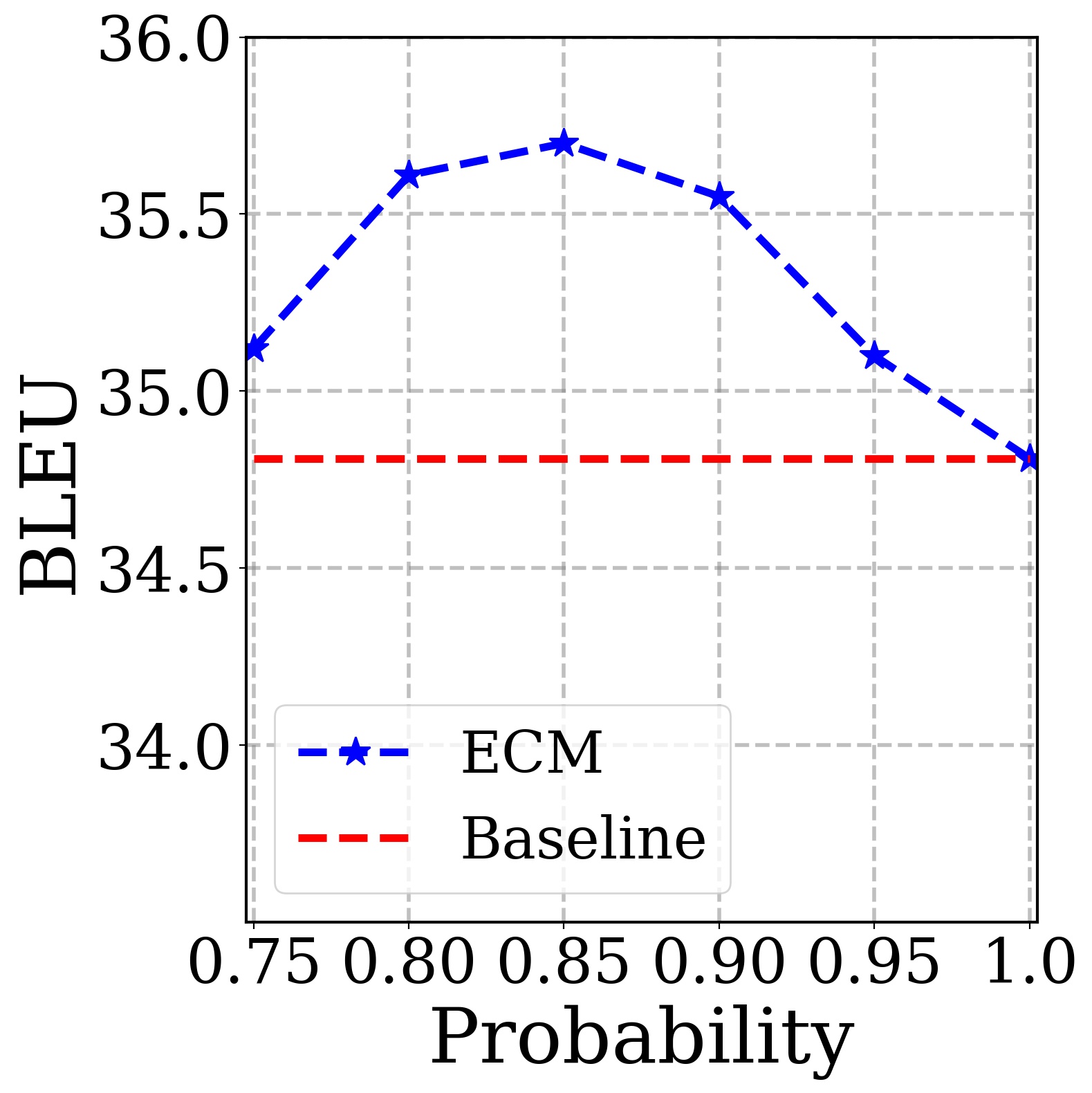}
        \caption{Study of $\beta$}
        \label{hyper_b}
    \end{subfigure}
    \caption{Results on IWSLT14 De$\to$En with different start steps $\alpha$ for token sampling and maximum sampling probabilities $\beta$.}
    \label{hyper_prob}
\end{figure}

\subsection{Study of Sampling Probability $p(\cdot)$}
In this subsection, we study how the hyperparameters $\alpha$ and $\beta$ in sampling probability $p(\cdot)$ will affect the model accuracy to what extent. We conduct experiments on IWSLT14 De$\to$En dataset to investigate when to start scheduled sampling ($\alpha$) and the best choice of maximum sampling probability ($\beta$). As can be seen from Figure~\ref{hyper_prob}, we have the following observations: 
\begin{itemize}
    \item From Figure~\ref{hyper_a}, it can be seen that starting to sample tokens before 10K steps results in worse accuracy than the baseline, which is consistent with the previous hypothesis as mentioned in Section~\ref{sec3_3} that sampling tokens too early will affect the accuracy. After 10K steps, our method gradually surpasses the baseline and achieves the best accuracy at 30K steps. After 30K steps, the model accuracy drops a little. In summary, $\alpha = 30K$ is an optimal choice to start token sampling. 
    \item From Figure~\ref{hyper_b}, the model accuracy decreases promptly, no matter when the maximum sampling probability is too big (\ie, the sampled tokens is close to the ground truths) or too small (\ie, the sampled tokens are close to the predicted tokens). This phenomenon is also consistent with our previous analysis in Section~\ref{sec3_3}. Therefore, we choose $\beta=0.85$ as the maximum sampling probability. 
\end{itemize}

\subsection{Cases Study}
To better understand the advantages of our method in correcting error tokens, we also prepare some translation cases in IWSLT14 De$\to$En, as shown in Table~\ref{de-en_cases}. It can be seen that the baseline result deviates the ground truth too much, which only predicts five correct tokens. When further adding scheduled sampling, the quality of translation can be improved, but still generates error tokens like ``\emph{like a filmmaker had requested}", which cause the following error prediction. Finally, when using error correction mechanism, we find that the translation produced by our method is closer to the target sequence with only tiny errors. In our translation sentence, although a mismatched token ``\emph{movie}" is predicted, our model can still correct the following predicted sequence like ``\emph{called for a change}". The quality of our translation also confirms the effectiveness of our model in correcting error information and alleviating error propagation.

\section{Conclusion}
In this paper, we incorporated a novel error correction mechanism into neural machine translation, which aims to solve the error propagation problem in sequence generation. Specifically, we introduce two-stream self-attention into neural machine translation, and further design a error correction mechanism based on two-stream self-attention, which is able to correct the previous predicted errors while generate the next token. Experimental results on three IWSLT tasks and two WMT tasks demonstrate our method outperforms previous methods including scheduled sampling, and alleviates the problem of error propagation effectively. In the future, we expect to apply our method on other sequence generation tasks, \eg, text summarization, unsupervised neural machine translation, and incorporate our error correction mechanism into other advanced structures. 

\section*{Acknowledgments}
This paper was supported by The National Key Research and Development Program of China under Grant 2017YFB1300205.

\bibliographystyle{named}
\bibliography{ijcai20}

\begin{thebibliography}{}

\bibitem[\protect\citeauthoryear{Bahdanau \bgroup \em et al.\egroup
  }{2014}]{bahdanau2014nmt}
Dzmitry Bahdanau, Kyunghyun Cho, and Yoshua Bengio.
\newblock Neural machine translation by jointly learning to align and
  translate.
\newblock {\em arXiv e-prints}, abs/1409.0473, September 2014.

\bibitem[\protect\citeauthoryear{Bengio \bgroup \em et al.\egroup
  }{2015}]{Bengio2015SS}
Samy Bengio, Oriol Vinyals, Navdeep Jaitly, and Noam Shazeer.
\newblock Scheduled sampling for sequence prediction with recurrent neural
  networks.
\newblock In {\em NIPS}, pages 1171--1179. 2015.

\bibitem[\protect\citeauthoryear{Cho \bgroup \em et al.\egroup
  }{2014}]{cho2014learning}
Kyunghyun Cho, Bart van Merri{\"e}nboer, Caglar Gulcehre, Dzmitry Bahdanau,
  Fethi Bougares, Holger Schwenk, and Yoshua Bengio.
\newblock Learning phrase representations using {RNN} encoder{--}decoder for
  statistical machine translation.
\newblock In {\em EMNLP}, oct 2014.

\bibitem[\protect\citeauthoryear{Devlin \bgroup \em et al.\egroup
  }{2019}]{devlin2019bert}
Jacob Devlin, Ming-Wei Chang, Kenton Lee, and Kristina Toutanova.
\newblock {BERT}: Pre-training of deep bidirectional transformers for language
  understanding.
\newblock In {\em NAACL}, pages 4171--4186, 2019.

\bibitem[\protect\citeauthoryear{Hassan \bgroup \em et al.\egroup
  }{2018}]{hassan2018achieving}
Hany Hassan, Anthony Aue, Chang Chen, Vishal Chowdhary, Jonathan Clark,
  Christian Federmann, Xuedong Huang, Marcin Junczys-Dowmunt, William Lewis,
  Mu~Li, et~al.
\newblock Achieving human parity on automatic chinese to english news
  translation.
\newblock {\em arXiv preprint arXiv:1803.05567}, 2018.

\bibitem[\protect\citeauthoryear{He \bgroup \em et al.\egroup
  }{2018}]{he2018layer_wise}
Tianyu He, Xu~Tan, Yingce Xia, Di~He, Tao Qin, Zhibo Chen, and Tie-Yan Liu.
\newblock Layer-wise coordination between encoder and decoder for neural
  machine translation.
\newblock In {\em NIPS}, pages 7944--7954. 2018.

\bibitem[\protect\citeauthoryear{Kingma and Ba}{2014}]{kingma2014method}
Diederik~P. Kingma and Jimmy Ba.
\newblock Adam: A method for stochastic optimization.
\newblock In {\em International Conference for Learning Representations}, 2014.

\bibitem[\protect\citeauthoryear{Ott \bgroup \em et al.\egroup
  }{2019}]{ott-etal-2019-fairseq}
Myle Ott, Sergey Edunov, Alexei Baevski, Angela Fan, Sam Gross, Nathan Ng,
  David Grangier, and Michael Auli.
\newblock fairseq: A fast, extensible toolkit for sequence modeling.
\newblock In {\em NAACL}, June 2019.

\bibitem[\protect\citeauthoryear{Radford \bgroup \em et al.\egroup
  }{2019}]{radford2019language}
Alec Radford, Jeffrey Wu, Rewon Child, David Luan, Dario Amodei, and Ilya
  Sutskever.
\newblock Language models are unsupervised multitask learners.
\newblock {\em OpenAI Blog}, 1(8), 2019.

\bibitem[\protect\citeauthoryear{Ranzato \bgroup \em et al.\egroup
  }{2016}]{Marc2016exposure}
Marc'Aurelio Ranzato, Sumit Chopra, Michael Auli, and Wojciech Zaremba.
\newblock Sequence level training with recurrent neural networks.
\newblock In {\em ICLR}, 2016.

\bibitem[\protect\citeauthoryear{Sennrich \bgroup \em et al.\egroup
  }{2016}]{sennrich2016bpe}
Rico Sennrich, Barry Haddow, and Alexandra Birch.
\newblock Neural machine translation of rare words with subword units.
\newblock In {\em ACL}, August 2016.

\bibitem[\protect\citeauthoryear{Song \bgroup \em et al.\egroup
  }{2018}]{song2018dpn}
Kaitao Song, Xu~Tan, Di~He, Jianfeng Lu, Tao Qin, and Tie-Yan Liu.
\newblock Double path networks for sequence to sequence learning.
\newblock In {\em COLING}, pages 3064--3074, August 2018.

\bibitem[\protect\citeauthoryear{Song \bgroup \em et al.\egroup
  }{2019}]{song2019mass}
Kaitao Song, Xu~Tan, Tao Qin, Jianfeng Lu, and Tie-Yan Liu.
\newblock Mass: Masked sequence to sequence pre-training for language
  generation.
\newblock In {\em ICML}, pages 5926--5936, 2019.

\bibitem[\protect\citeauthoryear{Sutskever \bgroup \em et al.\egroup
  }{2014}]{Sutskever2014seq2seq}
Ilya Sutskever, Oriol Vinyals, and Quoc~V Le.
\newblock Sequence to sequence learning with neural networks.
\newblock In {\em Advances in Neural Information Processing Systems 27}, pages
  3104--3112. 2014.

\bibitem[\protect\citeauthoryear{Tan \bgroup \em et al.\egroup
  }{2019}]{tan2019efficient}
Xu~Tan, Yingce Xia, Lijun Wu, and Tao Qin.
\newblock Efficient bidirectional neural machine translation.
\newblock {\em arXiv preprint arXiv:1908.09329}, 2019.

\bibitem[\protect\citeauthoryear{Vaswani \bgroup \em et al.\egroup
  }{2017}]{Vaswani2017transformer}
Ashish Vaswani, Noam Shazeer, Niki Parmar, Jakob Uszkoreit, Llion Jones,
  Aidan~N Gomez, \L~ukasz Kaiser, and Illia Polosukhin.
\newblock Attention is all you need.
\newblock In {\em NIPS}, pages 5998--6008, 2017.

\bibitem[\protect\citeauthoryear{Venkatraman \bgroup \em et al.\egroup
  }{2015}]{Venkatraman2015ImprovingMP}
Arun Venkatraman, Martial Hebert, and J.~Andrew Bagnell.
\newblock Improving multi-step prediction of learned time series models.
\newblock In {\em AAAI}, 2015.

\bibitem[\protect\citeauthoryear{Wu \bgroup \em et al.\egroup
  }{2018}]{wu2018beyond}
Lijun Wu, Xu~Tan, Di~He, Fei Tian, Tao Qin, Jianhuang Lai, and Tie-Yan Liu.
\newblock Beyond error propagation in neural machine translation:
  Characteristics of language also matter.
\newblock {\em arXiv preprint arXiv:1809.00120}, 2018.

\bibitem[\protect\citeauthoryear{Wu \bgroup \em et al.\egroup
  }{2019}]{wu2019beyond}
Lijun Wu, Xu~Tan, Tao Qin, Jianhuang Lai, and Tie-Yan Liu.
\newblock Beyond error propagation: Language branching also affects the
  accuracy of sequence generation.
\newblock {\em IEEE/ACM Transactions on Audio, Speech, and Language
  Processing}, 27(12):1868--1879, 2019.

\bibitem[\protect\citeauthoryear{Xia \bgroup \em et al.\egroup
  }{2019}]{Yingce2019tied}
Yingce Xia, Tianyu He, Xu~Tan, Fei Tian, Di~He, and Tao Qin.
\newblock Tied transformers: Neural machine translation with shared encoder and
  decoder.
\newblock In {\em AAAI}, pages 5466--5473, 2019.

\bibitem[\protect\citeauthoryear{Yang \bgroup \em et al.\egroup
  }{2019}]{Yang2019XLNet}
Zhilin Yang, Zihang Dai, Yiming Yang, Jaime Carbonell, Russ~R Salakhutdinov,
  and Quoc~V Le.
\newblock Xlnet: Generalized autoregressive pretraining for language
  understanding.
\newblock In {\em NIPS}, pages 5754--5764, 2019.

\bibitem[\protect\citeauthoryear{Zhang \bgroup \em et al.\egroup
  }{2019}]{zhang-etal-2019-bridging}
Wen Zhang, Yang Feng, Fandong Meng, Di~You, and Qun Liu.
\newblock Bridging the gap between training and inference for neural machine
  translation.
\newblock In {\em ACL}, pages 4334--4343, July 2019.

\end{thebibliography}

\end{document}